\documentclass[10pt,twocolumn,letterpaper]{article}

\usepackage{wacv}
\usepackage{times}
\usepackage{epsfig}
\usepackage{graphicx}
\usepackage{amsmath}
\usepackage{amssymb}
\usepackage{arydshln}
\usepackage{colortbl}
\usepackage{color}
\usepackage{float}
\usepackage{caption}
\usepackage{float}
\usepackage{multirow}
\usepackage{cite}

\wacvfinalcopy 

\def\wacvPaperID{****} 

\ifwacvfinal\fi
\begin{document}

\title{A Deep Four-Stream Siamese Convolutional Neural Network with Joint Verification and Identification Loss for Person Re-detection}

\author{Amena Khatun \hspace{2cm} Simon Denman \hspace{2cm} Sridha Sridharan \hspace{2cm} Clinton Fookes \\
Image and Video Laboratory,
Queensland University of Technology (QUT), Brisbane, QLD, Australia\\
{\tt\small Email: \{a2.khatun, s.denman, 
s.sridharan, c.fookes\}@qut.edu.au}
}


\maketitle
\ifwacvfinal\thispagestyle{empty}\fi

\begin{abstract}
 State-of-the-art person re-identification systems that employ a triplet based deep network suffer from a poor generalization capability. In this paper, we propose a four stream Siamese deep convolutional neural network for person re-detection that jointly optimises verification and identification losses over a four image input group. Specifically, the proposed method overcomes the weakness of the typical triplet formulation by using groups of four images featuring two matched (i.e. the same identity) and two mismatched images. This allows us to jointly increase the inter-class variations and reduce the intra-class variations in the learned feature space. The proposed approach also optimises over both the identification and verification losses, further minimising intra-class variation and maximising inter-class variation, improving overall performance. Extensive experiments on four challenging datasets, VIPeR, CUHK01, CUHK03 and PRID2011, demonstrates that the proposed approach achieves state-of-the-art performance.
\end{abstract}

\section{Introduction}

\begin{figure}
\begin{center}
\includegraphics[width=0.90\linewidth]{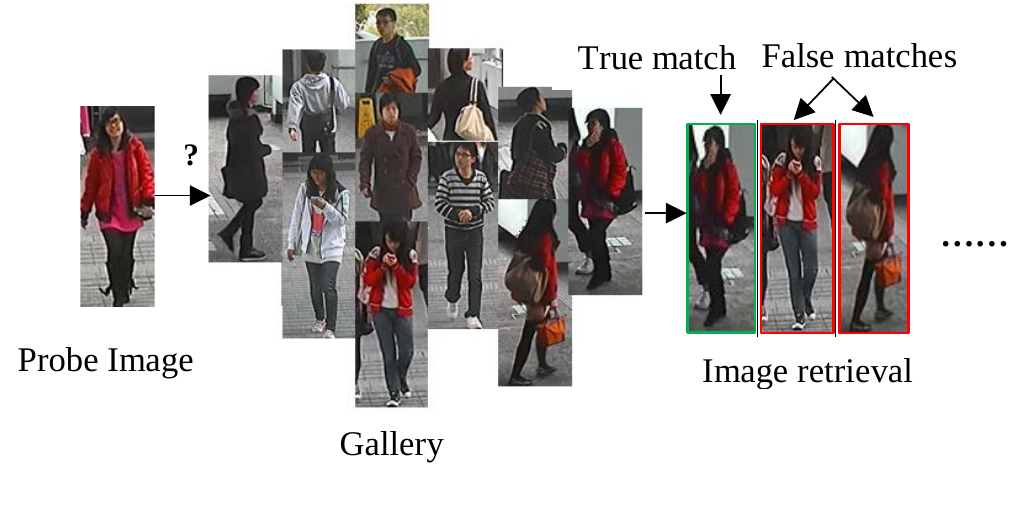}
\end{center}
   \caption{An end-to-end person re-detection system. A single probe image is compared to a large set of gallery images to locate other instances of the same subject.}
\label{fig:re}
\end{figure}

Person re-detection is the task of matching pedestrians across different spatial and temporal locations over multiple cameras (see Figure \ref{fig:re}). For instance, determining whether the person observed at point A is the same person now visible at point B. Despite the efforts of researchers, it remains
an unsolved problem, with the most challenging issue being to adapt to large changes in appearance caused by variations in lighting, and changes in the pose of the subject and the camera. Existing research on person re-detection has mainly focused extracting better features from the input images by developing feature extraction methods that are invariant to changes in condition; developing distance metrics to compare pairs of extracted features; or both.

Over the past few years, deep convolutional neural network (DCNN) methods have shown their potential with significant performance improvements in person re-detection. Some researchers \cite{6909421,7299016,7780513} have considered person re-detection as a multi class recognition problem and classification loss is used to train the network; while others consider it a verification problem \cite{DBLP:journals/corr/ChenGL15,Cheng_2016_CVPR,Ding:2015:DFL:2796563.2796623,7780513}. 

For verification, most researchers adopted either a Siamese network or a triplet based network. Siamese architectures take a doublet (pair of images) as input and pull the images of the same person close together in feature space, while features extracted from different people will be kept separate from each other in the feature space.

Usually, verification or similarity regression models take a pair of images as input for training purposes, which are then used to extract identifiable features to determine whether the query images (person-of-interest) are of the same person or not. On the contrary, identification or multi class recognition models classify an input image into a large number of identity
classes to predict the identity of the input image. Verification
and identification models are shown in Figure 2 and
Figure 3, respectively. With regards to feature extraction,
both approaches are different and have their own strengths
and weaknesses.

The two approaches also differ in that an identification
approach necessitates a closed-world view, while the verification
approach allows an open-world view. The identification
methods assume that the identities of subjects to be
recognized exist in the gallery, and subjects outside this list
cannot be identified. By contrast, a verification task considers
person re-detection as an open world scenario by using a
distance measure and threshold to determine if two images
belong to the same person or not. On the other hand, triplet based models takes three images as input: an anchor image, a positive image and a negative image; and the network enforces that the distance between the anchor and positive image should be less than the distance between the anchor and negative image. Essentially, the overall network architecture is a Siamese network with either two or three branches for the pairwise and triplet loss, respectively. However, the triplet loss forces that the distance of intra-class identity to be less than the distance of inter-class identities only in cases where the test images are from the same identity. In a real world scenario, test subjects and their images are totally unseen, thus the triplet based framework suffers from a poor generalization capability. 

\begin{figure}
\begin{center}
\includegraphics[width=0.65\linewidth]{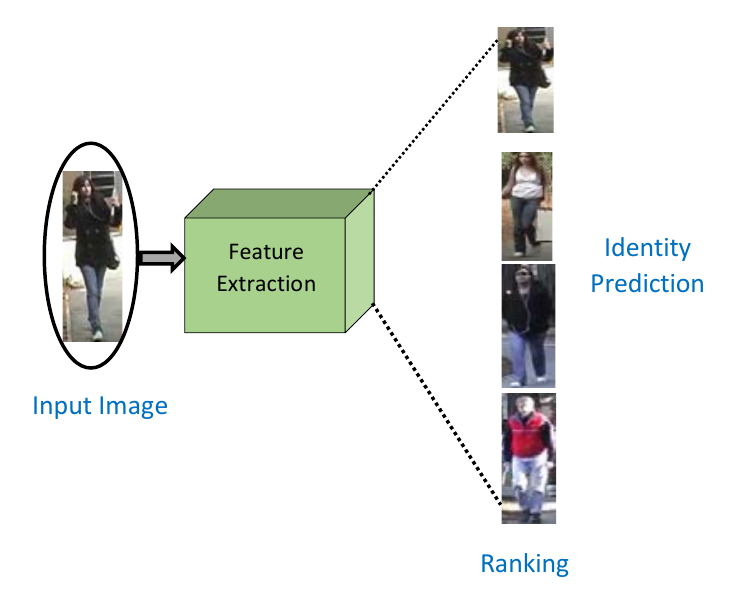}
\end{center}
   \caption{Identification based approach for person re-detection.
A single input image is mapped to a gallery to indicate the identity. This approach encourages images from different identities to be well separated in the feature space, though does not guarantee that images from the same person are close in the feature space.}
\label{fig:id}
\end{figure}

\begin{figure} 
\begin{center}
\includegraphics[width=0.65\linewidth]{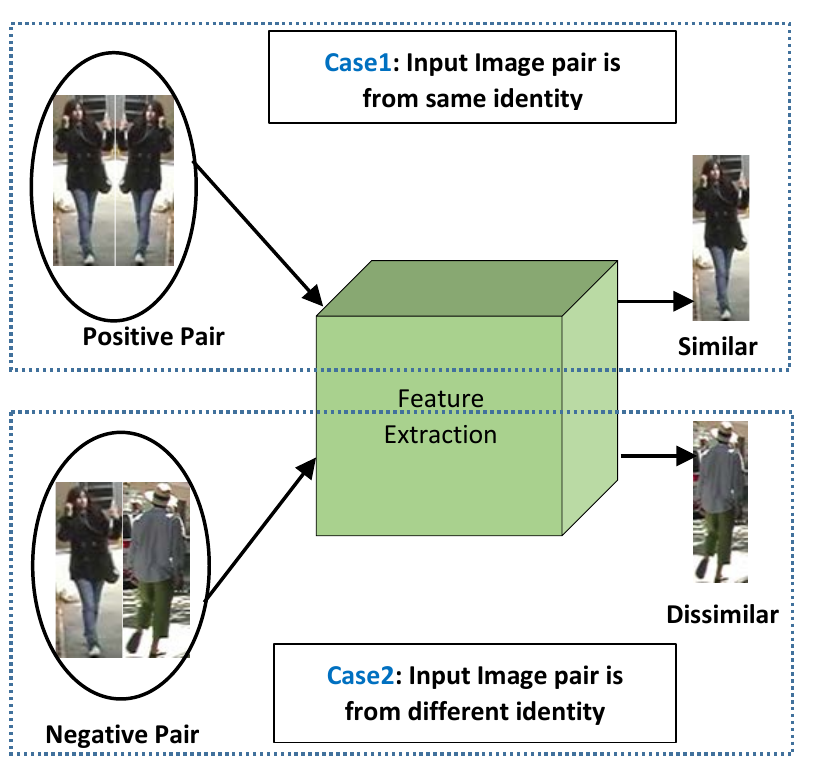}
\end{center}
\caption{Verification Model for person re-detection. Extracted
features are compared to determine if images belong
to the same person or not. In contrast to the identification
approach, the verification model promotes images of the
same person being close in feature space, though does not
ensure that images of different people are well separated.}
\label{fig:ve}
\end{figure}

The drawback of the verification model is that it does not
consider all the information in the training data. This
model considers only doublet (pairwise) or triplet labels of the relationship between a pair of images instead of other
images in the dataset \cite{DBLP:journals/corr/SunWT14}. An identification model overcomes this by extracting non-linear features directly from the input image to learn the person ID (identification) during training, and the loss function is used in the final layer makes full use of the re-ID labels. However, when used outside of training, to overcome the closed-world nature of the training approach, identification methods extract features from the last fully connected layer and compare these, however there is no guarantee that features from the same subject will lie close to one another in the feature space. As can be seen from the above, the two approaches have contrasting strengths where an identification approach will increase inter-class variations and a verification approach will minimise the intra-class variations. 

Motivated by the above observations, we propose to introduce a new loss function for person re-detection which we term the quartet loss function. This loss function enforces that the distance between the matched image pair is less than the distance between the mismatched image pair with respect to the same probe image, while simultaneously enforcing that the distance between positive pairs is less than the distance between negative pairs with respect to different probe images, resulting in better generalization and better performance. The proposed approach also combines elements of the verification and identification models to take full advantage of both approaches to maximize the performance of the person re-detection system. To summarize, our contributions are:

\begin{itemize}
\item We introduce a new loss function called the quartet loss to train the network where four images are taken as input, improving the generalization capability of the network, thus the trained network is able to differentiate between the positive and negative pairs not only with the same probe image, but also with respect to different probe images.

\item We propose a unified deep learning person re-detection network that jointly optimizes both identification (who is this person?) and verification tasks (are these images of the same person?) to maximize their strengths, thus improving person re-identification accuracy.

\item  We report a thorough experimental validation on the proposed losses in order to show how each performs separately i.e. only verification vs only identification vs identification and verification, to justify that the fusion of identification and verification helps to improve the performance.

\item We report state-of-the-art accuracy compared to existing methods on four challenging person re-detection datasets: VIPeR \cite{1478416}, CUHK01 \cite{6619305}, CUHK03 \cite{6909421} and PRID2011 \cite{conf/scia/HirzerBRB11}.

\end{itemize}
The rest of the paper is organized as follows: Section 2 outlines existing research relating to person re-detection; Section 3 describes our proposed methodology; Section 4 presents our experimental setup and results; and Section 5 concludes the paper.

\section{Related research}

For person re-detection, several hand-crafted feature methods such as colour histograms \cite{1640938,4270145,5995598,xiong2014person,Zhao_2013_ICCV,6619304,Liao_2015_CVPR} symmetry-driven accumulation of local features (SDALF) \cite{5539926}, adaboost \cite{1478416}, local binary pattern (LBP) \cite{6619305,6247939,Martinel2014SaliencyWF,xiong2014person}, gabor features \cite{6619305,6226421}, local maximum occurrence (LOMO) \cite{7410777,Liao_2015_CVPR}, speeded up robust features (SURF) \cite{BAY2008346}, spatiotemporal HOGs (STHOGs) \cite{6460721}, and scale-invariant feature transform (SIFT) \cite{Lowe2004} have been used for feature extraction. Once the features are extracted, a distance metric such as euclidean distance \cite{EURECOM+3274}, Bhattacharyya coefficients \cite{EURECOM+3274}, large margin nearest neighbour (LMNN) \cite{Weinberger:2009:DML:1577069.1577078}, relative distance comparison (RDC) \cite{6226421}, probabilistic relative distance comparison (PRDC) \cite{5995598}, local fisher discriminant analysis (LFDA) \cite{Sugiyama:2007:DRM:1248659.1248694,6619270}, Mahalanobis distance metric \cite{Roth_mahalanobisdistance}, or locally-adaptive decision function (LADF) \cite{Li_2013_CVPR} is considered to compare the extracted features for person re-identification. However, handcrafted features are not discriminative enough, and in particular they are not reliable and invariant against changes in pose, viewpoint, illumination and scene occlusion.

Deep convolutional neural networks (DCNNs) have shown tremendous potential in a variety of pattern recognition and machine learning tasks \cite{NIPS2012_4824,DBLP:journals/corr/RazavianASC14}. By learning feature representations directly from the data, DCNNs have demonstrated their power in complex and varied image and computer vision tasks, and have shown an ability to uncover complex relationships within the data. Some researchers considered DCNNs for person re-detection and the performance has been shown to be comparatively better than using handcrafted features.

The approaches of \cite{6909421,7299016,7780513,DBLP:journals/corr/VariorSLXW16,DBLP:journals/corr/VariorHW16,6976727} adopted a
Siamese architecture for person re-detection and treated the
task as a classification problem. The core concept of the
Siamese Convolutional Neural Network (SCNN) architecture
is to build a system where images belonging to the
same identity will be close to each other in the feature
space, whereas images from different identities will well
separated. Although \cite{6976727} first proposed the Siamese based
classification model for person re-detection, \cite{6909421} matched the filter responses of local patches of images by introducing a filter pairing neural network and \cite{7299016} presented a new layer to compute the neighbourhood difference between two input images. Varior et al. \cite{DBLP:journals/corr/VariorHW16} introduced a gating function with the Siamese architecture to capture the subtle structures of a pair of probe images. This gate function was used after each convolution layer. The limitation of this method is that the probe image has to be paired with each gallery image before being fed into the network which is inefficient when dealing with large datasets. Similar to \cite{DBLP:journals/corr/VariorHW16}, \cite{DBLP:journals/corr/LiuFQJY16} proposed a Siamese architecture and adopted a soft attention based model to focus on the local body parts of an input image pair instead of global body parts. Another study \cite{McLaughlin_2016_CVPR} introduced a video-based re-detection system where the SCNN was used for feature extraction followed by a recurrent neural network (RNN) layer to transfer information within time-steps. In \cite{Wang_2014_CVPR}, triplet units were adopted to train the network with the triplet loss function in order to learn image similarity metrics. In \cite{Ding:2015:DFL:2796563.2796623}, a triplet network was used for person re-detection where only ranking was taken into consideration. The triplet loss function was improved in \cite{Cheng_2016_CVPR} by inserting a new term within the original triplet loss function and achieved state-of-the-art performance, however they considered person re-detection as a verification task. In \cite{DBLP:journals/corr/ChenCZH17}, a new loss function was introduced where a four image input is taken into consideration for training, where two margins (similar to \cite{Cheng_2016_CVPR}) are used. The function of the first margin is the same as the traditional triplet loss while the second margin is used to further maximize the inter-class distance. As the second margin is weaker than the first margin, in their ranking loss the triplet loss dominates which forces to the network minimize the intra-class distance only in cases where the test images are from the same identity. By contrast, in our proposed loss, we equally consider minimizing the intra-class distance in the case of either having the same or different probe images. Furthermore, \cite{DBLP:journals/corr/ChenCZH17} considered person re-detection as a verification task  while we consider it jointly as a verification and identification task. 

Form the above analysis, it is noted that most of the research adopted either pairwise or triplet verification loss for person re-detection, although the overall network is typically a Siamese based architecture. Although \cite{AAAI1714313} considered person re-detection jointly as a ranking and classification problem, they adopted the traditional triplet loss which forces that the distance of intra-class identities to be less than the distance of inter-class identities only in cases where the test images are from the same identity. Wang et al. \cite{7780513} also considered person re-detection as both verification and identification tasks, however, they trained the two losses separately and fuse these two tasks only at the score level. Thus the trained network is not be able to learn to perform both tasks in a single framework.

Inspired by the success of deep convolutional neural networks for person re-detection, we propose a deep four stream convolutional architecture that differs from the above deep learning based methods in network architecture and loss function. Specifically, we introduce a unified deep network with quartet loss that simultaneously optimises for the identification and verification tasks in order to re-detect the target person. By taking a group of four images as a training unit we can achieve better re-detection performance as the network is better equipped to increase the inter-class distance while decreasing the intra-class distance with respect to both the same probe image (similar to the original triplet loss) and a different probe image (our proposed quartet loss) simultaneously. Furthermore, we provide a comparison of the joint verification and identification tasks with only verification and only identification to show that learning the tasks jointly performs better than either one alone.

\section{Proposed Methodology}
In this paper, we consider person re-detection jointly as an identification and verification task, in contrast to other approaches which consider it either as a multi class recognition (identification) task or a similarity regression (verification) task. The verification task considers person re-detection as a binary class recognition problem which determines whether the probe image belongs to the same person or not while the identification model is a multi-class recognition task, which predicts the identity of the probe image. These two tasks are combined in our model to re-detect people with increased accuracy.

\subsection{The Overall Framework}
Our network is a four stream Siamese convolutional neural network with a quartet loss function that combines the verification and identification losses. We use a group of four images as input for training purposes as illustrated in Figure 4, followed by a quartet loss and a softmax loss function for verification and identification tasks,  respectively. In this paper, the input selection is denoted as, $A_i$ = $A_i^1$, $A_i^2$, $A_i^3$, $A_i^4$ where $A_i^1$ is the anchor image, $A_i^2$ is the positive image (i.e. $A_i^1$ and $A_i^2$ are the images belonging to the same person) while $A_i^3$ and $A_i^4$ are the negative images (i.e. images from two different people, both different to the person in the positive images). The network consists of four ImageNet \cite{NIPS2012_4824} pre-trained CNN models and two losses. The four images share the weights and biases through the four CNNs. After training, the CNN architecture is able to differentiate the matched and mismatched pairs through the learned feature space with respect to the same and different probe images. In the model, Alexnet is used as a pre-trained network, thus we follow the same architecture as Alexnet, consisting of five convolution layers and three fully connected layers. From \cite{7410714}, the lower layers of deep architectures encode more discriminative features to capture intra-class variations and provide more detailed local features, whereas higher layers capture semantic features. Thus for verification, we compare the images based on low level features. For identification, features are extracted from higher layers as these features focus on identifiable local semantic concepts.

\begin{figure*}
\begin{center}
\includegraphics[width=0.8\linewidth]{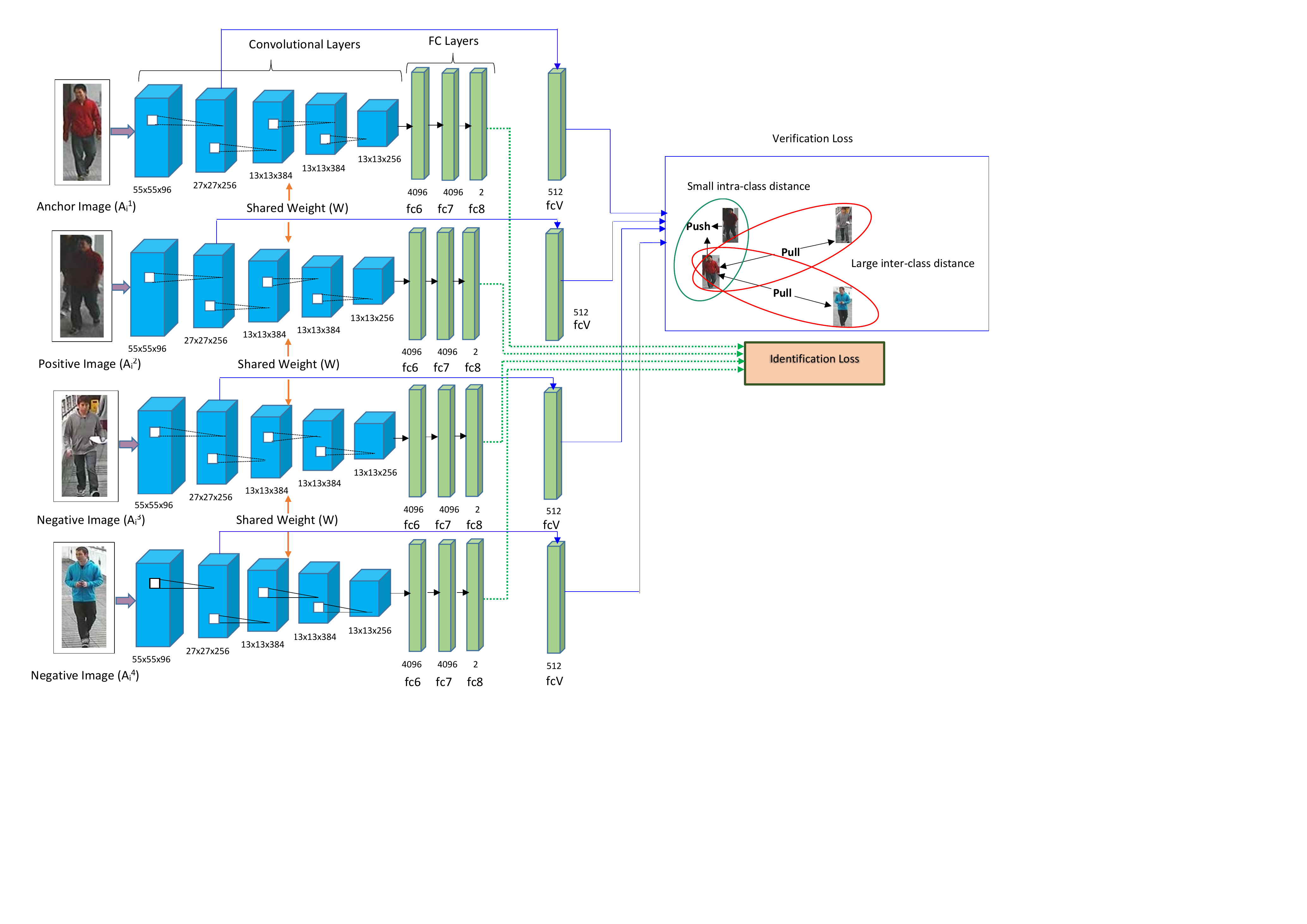}
\end{center}
\caption{The overall architecture of the proposed model. Given four images of size 227 x 227 as input, four identical Alexnet models sharing weights and biases are used to extract features in a identical way. Then, the fcV features from the four streams of the CNN are used to predict the verification label and fc8 features are used to identity of the four input images respectively.}
\label{fig:me}
\end{figure*}

\subsubsection{Four Stream Siamese Based Verification and Identification Model with Quartet loss}
Our proposed quartet loss is designed based on the triplet loss, thus, in this section we introduce the original triplet loss and then we present our proposed loss.

In the triplet network, three images are used as an input to train the network where for each positive pair, a third (negative) image is randomly selected to complete the group of three (i.e. input triplet $A_i$ = $A_i^1$, $A_i^2$, $A_i^3$, $A_i^1$ and $A_i^2$ are images from the same person and $A_i^1$ and $A_i^3$ belong to a different person). The learned feature space is expressed as, $\Theta_w(I)$ which is the feature representation for image $I$. The learned feature satisfies the following condition under Euclidean distance or $L_2$ norm,
\begin{equation}\parallel \Theta_w (A_i^1) - \Theta_w (A_i^2) \parallel^2 < \parallel \Theta_w (A_i^1) - \Theta_w (A_i^3) \parallel^2.
\end{equation}
In other words, in the learned feature space the distance between a negative pair of images should be greater than the distance between a positive pair. If three images are taken as input, the triplet loss function can be formulated as,
\begin{align}
Loss_{triplet} = \sum_{i=1}^{n} \Big(max\big\{\parallel\Theta_w (A_i^1) -  \Theta_w (A_i^2) \parallel^2 \nonumber \\ - \parallel \Theta_w (A_i^1) - \Theta_w (A_i^3) \parallel^2,   margin\big\}\Big).
\end{align}
Thus, the triplet loss enforces that the distance between the matched pair should be smaller than the distance between the mismatched pair by a predefined margin which enables the network to differentiate positive and negative samples when the probe image belongs to the same person. The max operation is used with the margin to prevent the overall value of the objective function from being dominated by easily identifiable triplets, the same technique as in hinge-loss functions.

In our proposed quartet network, four images are taken as an input to train the network where for each positive pair, third and fourth (negative) images are randomly selected to complete the group of four (i.e. input unit $A_i$ = $A_i^1$, $A_i^2$, $A_i^3$, $A_i^4$, $A_i^1$ and $A_i^2$ are images from the same person and $A_i^3$ and $A_i^4$ belong to different people). All the images are resized to 227 x 227 and fed into four DCNNs which share the same weights and biases, thus the feature extraction will be consistent for each image. The proposed quartet loss can be expressed as,
\begin{align}
Loss_{quartet} = \sum_{i=1}^{n} \Big(max\big\{\parallel\Theta_w (A_i^1) - \Theta_w (A_i^2) \parallel^2 \nonumber \\  - \parallel \Theta_w (A_i^1) - \Theta_w (A_i^3) \parallel^2 + \parallel\Theta_w (A_i^1) - \Theta_w (A_i^2)\parallel^2\nonumber\\-\parallel\Theta_w (A_i^4) - \Theta_w (A_i^3) \parallel^2, margin\big\}\Big).
\end{align}

In other words, the quartet loss increases the inter-class distance over the intra-class distance with the same target image as well as different target images, improving the overall performance. The positive pair is comprised of $\Theta_w (A_i^1)$ and $\Theta_w (A_i^2)$ and is included twice, to compensate for having two negative pairs (i.e. $\Theta_w (A_i^1)$, $\Theta_w (A_i^3)$; and $\Theta_w (A_i^3)$, $\Theta_w (A_i^4)$). The first of the negative pairs shares a common probe image with the positive pair (i.e. $\Theta_w (A_i^1)$), while the second negative pair uses two different images. Thus, the proposed verification loss is able to maximize  the inter-class distance even if the target image comes from a different identity. As the four image input is incremental as the dataset gets larger, a margin is set to restrict the distance between positive and negative pairs.
Here the square of the distance is used to simplify the partial derivative calculation.  Gradient descent is used to train the deep convolutional network and the gradient is calculated as partial derivatives of Equation (3) using the chain rule. The gradient is obtained as follows,
\begin{align}
\frac{\partial d(A_i, w)}{\partial w }=2(\Theta_w (A_i^1)-\Theta_w (A_i^2)) \frac{\partial \Theta_w (A_i^1)-\partial \Theta_w (A_i^2)}{\partial w} \nonumber  \\ - 2(\Theta_w (A_i^1)-\Theta_w (A_i^3)) \frac{\partial \Theta_w (A_i^1)-\partial \Theta_w (A_i^3)}{\partial w} \nonumber \\ +2(\Theta_w (A_i^1)-\Theta_w (A_i^2)) \frac{\partial \Theta_w (A_i^1)-\partial \Theta_w (A_i^2)}{\partial w} \nonumber  \\ - 2(\Theta_w (A_i^4)-\Theta_w (A_i^3)) \frac{\partial \Theta_w (A_i^4)-\partial \Theta_w (A_i^3)}{\partial w}.
\end{align}

The values for Equation (4) can be acquired by forward and back propagation for each input image in the training input unit. We note that not only does this formulation help improve learning, but the use of quartets helps generate additional training permutations of the data to actually train the network, helping to overcome the relatively small amount of training data typically available in this domain. 

In the identification portion, to determine the identity of the probe images the network enforces images of different identities to be far apart in the learned feature space. As our training network consists of four images as an input, this leads to three pairs being formed for identification (one from images of the same identity and two from images of different identity). The identification loss is determined using a softmax layer to classify the similarity between the test images and the gallery images. Similar to traditional multi-class recognition approaches, we use the cross-entropy loss for identity prediction, which is defined as,
\begin{equation}
Loss_{identification}= - \sum_{i=1}^{n}p_ilog\bar{p_i}=-log\bar{p}_t,
\end{equation}
where $p_i$ is the probability distribution of the target, $i$ is the number of classes, $\bar{p}_i$ is the predicted probability distribution and  $t$ is the target class. In our case, we can define it simply as a binary cross-entropy loss as this is a cross entropy for the two-class case. Thus, the output will be 1 if the probe image is matched with the positive pair otherwise it will be 0 in case of a mismatched pair.

\subsection{Network Architecture}
Our network consists of five convolution layers as shown in Figure 4: the first two are used for verification and identification; and the remaining thre are used only for identification. For verification, features extracted from the first two convolutional layers are sent to a fully connected layer followed by the verification loss function. For the identification task, after the two convolutional layers, three pairs are formed as mentioned above (one comprising images of the same identify, two being image pairs with different identifies) which are sent to the three convolutional layers followed by a softmax layer to determine the similarity. In the test phase, the two input images are sent through all layers, where the last layer obtains the similarity probability of a test pair. These similarity scores are used to rank probe subjects similarly to each gallery image.

\section{Experiments \& Results}
We conduct extensive experiments to evaluate the performance of the proposed method on four person re-detection datasets: VIPeR, CUHK03, CUHK01 and PRID2011.

\subsection{Datasets}
We used four challenging and popular person re-detection benchmark datasets: VIPeR, CUHK03, CUHK01, and PRID2011; for our experiments. All the datasets have images captured by different cameras. A brief explanation on four datasets are as follows:

\begin{itemize}
\item VIPeR \cite{1478416} dataset: Consists of 632 identities, each of which has a pair of images captured with different cameras with distinct angles, illuminations and poses. In our experiment, 316 identities are randomly chosen for testing, leaving the rest for training, as per \cite{Cheng_2016_CVPR} .

\item CUHK03 \cite{6909421} dataset: Consists of 14097 images of 1467 identities collected from 6 surveillance cameras in the CUHK campus. In this dataset, each identity is taken from two disjoint camera views. We use the training and testing sets as proposed in \cite{zhao2017spindle}.

\item CUHK01 \cite{6619305} dataset: Consists of 971 persons from two camera views. Each person has four images, two from each camera. We use the training and testing sets as proposed in \cite{zhao2017spindle}.

\item PRID2011 \cite{conf/scia/HirzerBRB11} dataset: The PRID2011 dataset extracts images from video recorded by static surveillance cameras with two camera views, each of which contains 385 and 749 identities. In this dataset, only 200 persons are seen in both camera views. We follow the same training and testing sets as \cite{DBLP:journals/corr/XiaoLOW16}.

\end{itemize}

\subsection {Evaluation Protocol}
We use cumulative matching characteristic (CMC) curves for performance evaluation, which are widely used for person re-detection evaluations, and determine rank-1, rank-5, and rank-10 accuracy. The CMC represents the probability of finding the correct match for the probe image in the top n matches where the best case is given by a re-detection rate of 100\% at the rank-1, i.e. the correct match is always the highest rank. CMC can be defined as:
\begin{equation}
CMC (i)= \sum_{r=1}^{i}q(r),
\end{equation}
where, $i$ represents the rank and $q(r)$ is the number of correct re-detected queries.

\subsection{Experimental Setup }

The proposed method is implemented using the Caffe framework \cite{Jia:2014:CCA:2647868.2654889}. All images are resized to 227x227 for Alexnet during training. The network is pre-trained on ImageNet \cite{NIPS2012_4824}. We set the learning rate to 0.0001 throughout the experiments. Batch size is set to 128 and we train the network for 30,000 iterations with an NVIDIA GPU, which takes about 24-30 hours. We use stochastic gradient descent (SGD) to update the parameters of the network. Our results on four different datasets (VIPeR, CUHK03, CUHK01, and PRID2011) are given in Tables \ref{tab:viper} to \ref{tab:prid2011} respectively, alongside other state-of-the-art approaches on these datasets.

\subsection {Comparisons with state-of-the-art approaches}
As shown in Table \ref{tab:viper}, \ref{tab:cuhk03}, \ref{tab:cuhk01} and \ref{tab:prid2011}, we compare the results of our proposed method with state-of-the-art approaches on four datasets (VIPeR, CUHK03, CUHK01, and PRID2011) in terms of rank-1, rank-5 and rank-10 accuracy. We report the single-query evaluation results. We also report the result of the proposed approach using only verification and only identification losses, which is compared with  the proposed joint verification and identification task.

\begin{table}[!htbp]
\fontsize{8}{9}\selectfont 
\begin{center}
\begin{tabular}{l||c|c|c|c|c|c|c||cc}
\hline
\textbf{Method} & \textbf{Type} & \textbf{Rank1} & \textbf{Rank5} & \textbf{Rank10}\\
\hline\hline
DML \cite{6976727} & ID &28.23 &59.27 &73.45 \\
FLCA \cite{7900000} & ID &42.5 &72 &91.72\\
DGD \cite{DBLP:journals/corr/XiaoLOW16} &ID   &38.6 &- &-\\
Improved Trp \cite{Cheng_2016_CVPR} &V &47.8 &74.7 &84.8\\
SIRCIR \cite{7780513} & ID+V &35.76 &- &- \\
GSCNN \cite{DBLP:journals/corr/VariorHW16} & ID &37.8 &66.9 &77.4\\
DRPR \cite{DBLP:journals/corr/ChenGL15} &V &38.37 &69.22 &81.33\\
DRDC \cite{Ding:2015:DFL:2796563.2796623} &V &40.5 &60.8 &70.4\\
IDLA \cite{7299016} & ID &34.81 &63.32 &74.79\\
EDM \cite{Shi2016} & ID &40.91 &- &-\\
MTDNet \cite{AAAI1714313} & ID+V &45.89 &71.84 &83.23\\
BTL \cite{DBLP:journals/corr/ChenCZH17} & V &49.05 &73.10 &81.96 \\
Spindle Net \cite{zhao2017spindle} &ID &53.8 &74.1 &83.2\\
\hline
\textbf{The proposed method} & V & 62.5 & 83.1 &92.4\\
\textbf{The proposed method} & ID & 58.2 & 76.7 & 85.8\\
\textbf{The proposed method} & \textbf{ID+V} & \textbf{68.7} & \textbf{88.9} & \textbf{94.6}\\
\hline
\end{tabular}
\end{center}
\caption{Results of our proposed method in comparison with other state-of-the-art approaches on the VIPeR \cite{1478416} dataset.  In the type column, ID indicates an identification loss is used, V that a verification loss is used, and ID + V indicates that a combined loss is used.}
\label{tab:viper}
\end{table}

\begin{table}[!htbp]
\fontsize{8}{9}\selectfont 
\begin{center}
\begin{tabular}{l||c|c|c|c|c|c|c||cc}
\hline
\textbf{Method} & \textbf{Type} & \textbf{Rank1} & \textbf{Rank5} & \textbf{Rank10}\\
\hline\hline
PersonNet \cite{DBLP:journals/corr/WuSH16}  &V &64.8 &89.4 &94.92 \\
FPNN \cite{6909421} & ID &20.65 &51.32 &68.74\\
CAN \cite{DBLP:journals/corr/LiuFQJY16} & ID &72.3 &93.8 &98.4 \\
DGD \cite{DBLP:journals/corr/XiaoLOW16} &ID   &75.3 &- &-\\
SIRCIR \cite{7780513} & ID+V &52.17 &- &- \\
DLCNN \cite{DBLP:journals/corr/ZhengZY16} & ID+V & 83.4 &97.1 &98.7 \\
GSCNN \cite{DBLP:journals/corr/VariorHW16} & ID &68.1 &88.1 &94.6\\
IDLA \cite{7299016} & ID &54.74 &- &-\\
EDM \cite{Shi2016} & ID &61.32 &- &-\\
MTDNet \cite{AAAI1714313} & ID+V &74.68 &95.9 &97.47\\
BTL \cite{DBLP:journals/corr/ChenCZH17} & V &75.53 &95.15 &99.16 \\
Spindle Net \cite{zhao2017spindle} &ID & \textbf{88.5} &97.8 &98.6\\
\hline
\textbf{The proposed method} & {V} &81.14  &95.6  &98.7 \\
\textbf{The proposed method} & {ID} & 76.9& 91.3 &97.4\\
\textbf{The proposed method} & \textbf{ID+V} & {85.5} & \textbf{98.74} & \textbf{99.8}\\
\hline
\end{tabular}
\end{center}
\caption{Results of our proposed method in comparison with other state-of-the-art approaches on CUHK03 \cite{6909421} dataset. Type column is as per Table 1.}
\label{tab:cuhk03}
\end{table}

\begin{table}[!htbp]
\fontsize{8}{9}\selectfont 
\begin{center}
\begin{tabular}{l||c|c|c|c|c|c||cc}
\hline
\textbf{Method} & \textbf{Type} & \textbf{Rank1} & \textbf{Rank5} & \textbf{Rank10}\\
\hline\hline
FPNN \cite{6909421} & ID &27.87 &- &-\\
FLCA \cite{7900000} & ID &46.8 &71.8 &80.5\\
DGD \cite{DBLP:journals/corr/XiaoLOW16} &ID    &71.7 &88.6 &92.6\\
Improved Trp \cite{Cheng_2016_CVPR} &V &53.7 &84.3 &91\\
SIRCIR \cite{7780513} & ID+V &71.8 &- &- \\
DRPR \cite{DBLP:journals/corr/ChenGL15} &V &70.94 &92.3 & {96.9}\\
IDLA \cite{7299016} & ID &65 &89.5 &93\\
PersonNet \cite{DBLP:journals/corr/WuSH16} &V &71.14 &90 & 95 \\
EDM \cite{Shi2016} & ID &69.38 &- &-\\
MTDNet \cite{AAAI1714313} & ID+V &77.5 &95.0 &97.5\\
BTL \cite{DBLP:journals/corr/ChenCZH17} & V &62.55 &83.44 &89.71 \\
Spindle Net \cite{zhao2017spindle} &ID &79.9 &94.4 &97.1\\
\hline
\textbf{The proposed method} & {V} & 80.12 &95.0 &97.23\\
\textbf{The proposed method} &ID &78.01 &93.4 &96.8\\
\textbf{The proposed method} & \textbf{ID+V}  &\textbf{83.95} & \textbf{98.15} & \textbf{98.97}\\
\hline
\end{tabular}
\end{center}
\caption{Results of our proposed method in comparison with other state-of-the-art approaches on CUHK01 \cite{6619305} dataset. Type column is as per Table 1.}
\label{tab:cuhk01}
\end{table}

\begin{table}[!htbp]
\fontsize{8}{9}\selectfont 
\begin{center}
\begin{tabular}{l||c|c|c|c|c|c||cc}
\hline
\textbf{Method} & \textbf{Type} & \textbf{Rank1} & \textbf{Rank5} & \textbf{Rank10}\\
\hline\hline
DML \cite{6976727} & ID &17.9 &37.5 &45.9 \\
DGD \cite{DBLP:journals/corr/XiaoLOW16} &ID &64 &- &-\\
Improved Trp \cite{Cheng_2016_CVPR} &V &22 &- &47\\
MTDNet \cite{AAAI1714313} & ID+V &32.0 &51.0 &62.0\\
\hline
\textbf{The proposed method} & V & 69 & 90 & 95\\
\textbf{The proposed method} & ID & 66 & 87 & 91\\
\textbf{The proposed method} & \textbf{ID+V} &\textbf{75} & \textbf{93} & \textbf{97}\\
\hline
\end{tabular}
\end{center}
\caption{Results of our proposed method in comparison with other state-of-the-art approaches on PRID2011 \cite{conf/scia/HirzerBRB11} dataset. Type column is as per Table 1.}
\label{tab:prid2011}
\end{table}

For the VIPeR dataset, our model achieves 68.7\% rank-1 accuracy, outperforming state-of-the-art approaches, \cite{Cheng_2016_CVPR} and \cite{zhao2017spindle} by 20.9\% and 14.9\%, respectively. Cheng et al. \cite{Cheng_2016_CVPR} considered person re-detection as a verification task and inserted a new term within original triplet loss, and \cite{zhao2017spindle} adopted identification task for person re-detection and extracted features from different body regions.

For the CUHK03 dataset, the largest dataset for person re-detection consisting of 1467 identities, our proposed method achieves 85.5\% rank-1 accuracy which outperforms most state-of-the-art approaches, such as, DLCNN \cite{DBLP:journals/corr/ZhengZY16} by 2.1\% who considered person re-detection both as a verification and identification task. But in contrast to our proposed work they adopted a pairwise verification loss whereas we consider a quartet verification loss, thus improving the generalization capability. Although in terms of rank-1 accuracy, \cite{zhao2017spindle} beats our result by 3\%, they considered human landmark information and were thus able to leverage more detailed information by dividing the body into seven regions for feature extraction. Thus with a larger dataset, they can extract more identifiable features from different regions. Despite this additional information however, our proposed method outperforms their method at rank-5 and rank-10 by 0.94\% and 1.20\% respectively.

For the CUHK01 dataset, the previous best methods are \cite{7780513}, \cite{DBLP:journals/corr/XiaoLOW16} and \cite{zhao2017spindle} with 71.8\% and 71.7\% and 79.9\% rank 1 accuracy respectively. In \cite{7780513}, both verification and identification tasks were adopted, but in contrast to our approach they simply join these two methods at the score level, whereas we combine the verification and identification tasks within the network architecture. In \cite{DBLP:journals/corr/XiaoLOW16}, Xiao et al. considered person re-detection as an Identification task where 6 datasets (VIPeR, CUHK01, PRID2011, i-LIDS, CUHK03, 3DPeS) were joined together for training purpose and \cite{zhao2017spindle} who extracted local features from seven body parts for re-detection. Notably, our rank-1 accuracy is 83.95\% which is much better than above state-of-the-art methods.

For the PRID2011 dataset, the existing best result was achieved by \cite{DBLP:journals/corr/XiaoLOW16}, which as noted above used 6 datasets for training and considered person re-detection as an identification task. By contrast, our combined verification and identification model with the four stream Siamese network performs better than this and other approaches, achieving a 75\% rank-1 accuracy.

From the above comparisons with other methods, it is worth noting that almost all the evaluated methods considered person re-detection either as an identification or verification problem alone except \cite{7780513} and \cite{AAAI1714313} who considered both tasks jointly, but \cite{7780513} combined them at the score level and \cite{AAAI1714313} adopted the traditional triplet loss function. In contrast, we fuse both tasks within the proposed network architecture, enabling the network to jointly learn to perform both with quartet loss function which can overcome the weakness of the triplet loss function by improving the generalization capability of the network. For completeness, we also evaluate the proposed approach using each loss on it's own. From Tables 1 to 4, we can clearly see that the proposed approach outperforms the state of the art in all but one situation (compared to [44] on CUHK-03 at Rank-1), and the joint loss clearly outperforms either loss individually, clearly indicating that person re-detection performs better if it is considered both as a verification and identification task rather than considering it as either one independently. Furthermore, we propose a quartet loss method that improves the generalization capability by increasing the inter-class variation with reducing the intra-class variation through the use of different probe images. This is evidenced by the highly competitive results of the proposed approach with only a single loss. For instance in Table 1 for the VIPeR dataset, using only the verification or identification loss outperforms all baselines at Rank-1, 5 and 10.

\subsection{Comparison Between Triplet and Quartet Losses}

\begin{table}[!htbp]
\fontsize{8}{8}\selectfont
\begin{center}
\begin{tabular}{l|c||c|c|c|}
\hline
\textbf{Database} & \textbf{Loss} & \textbf{Rank1} & \textbf{Rank5} & \textbf{Rank10}\\
\hline
\hline
\multirow{2}{*}{VIPeR} & Triplet & 61 & \textbf{92} & \textbf{95} \\
 & Quartet & \textbf{68.7} & 88.9 & 94.6 \\
\hline
\multirow{2}{*}{CUHK03} & Triplet & 83.8 & 97.8 & 98.8 \\
 & Quartet & \textbf{85.5} & \textbf{98.74} & \textbf{99.8} \\
\hline
\multirow{2}{*}{CUHK01} & Triplet & 79.5 &  93 & 94.5 \\
 & Quartet & \textbf{83.95} & \textbf{98.15} & \textbf{98.97} \\
\hline
\multirow{2}{*}{PRID2011} & Triplet & 71 & 91 & 96 \\
 & Quartet & \textbf{75} & \textbf{93} & \textbf{97} \\
\hline
\end{tabular}
\end{center}
\caption{Results of our proposed method in comparison with other state-of-the-art approaches on VIPeR \cite{1478416}, CUHK03 \cite{6909421}, CUHK01 \cite{6619305} and PRID2011 \cite{conf/scia/HirzerBRB11} dataset.}
\label{tab:tripletQuartet}
\end{table}

To evaluate the benefit of the quartet over the conventional triplet loss, we formulate the proposed joint verification and identification approach using a triplet of input images, and compare this to the complete proposed approach (quartet with joint verification and identification losses). Results are shown in Table \ref{tab:tripletQuartet}. From this, we can clearly see that the proposed use of a quartet of input images leads to improved performance. Furthermore, comparing Table \ref{tab:tripletQuartet} to Tables \ref{tab:viper}, \ref{tab:cuhk03}, \ref{tab:cuhk01} and \ref{tab:prid2011}, we can see that using the proposed dual losses with a triplet formulation is still able to outperform all other baselines on VIPeR and PRID2011, and is only bested by \cite{zhao2017spindle} on CUHK03 and CUHK01. This highlights the twin gains by using both the quartet of input images and the joint verification and identification losses.

\vspace{-2mm}
\section {Conclusion and Future work}
In this paper, we propose a deep learning person re-detection method by joint identification and verification learning. Since identification is adept at predicting the identity of the target image, and verification is effective in determining the relationship between target and gallery images, we combine these two approaches together to maximize the advantages of both of these methods. We further extend existing architectures through the use of a quartet of input images which are incorporated into the verification and identification loss function. This enforces the intra-class distance to be larger than the inter-class distance with respect to multiple different probe images. Experimental results shows the performance of the joint identification and verification method outperforms state-of-the-art techniques on four datasets, VIPeR, CUHK03, CUHK01, and PRID2011. In the future, we will investigate other ways to improve the performance of person re-identification, such as dividing the human body into local body parts for better feature extraction. Furthermore, we will expand our work by adding deep transfer learning to more effectively adapt to unseen conditions. Finally, more recent architectures such as ResNet \cite{He_2016_CVPR} will be investigated as an alternative feature extraction pipeline.\\

\vspace{-2mm}
\section*{Acknowledgement}
This research was supported by the Australian Research Council's Linkage Project ``Improving Productivity and Efficiency of Australian Airports'' (140100282). The authors would also like to thank QUT High Performance Computing (HPC) for providing the computational resources for this research.

{\small
\bibliographystyle{ieee}
\bibliography{egpaper_final}
}

\end{document}